\documentclass{article}

\PassOptionsToPackage{square, numbers}{natbib}

\usepackage[preprint]{neurips_2024}

\usepackage[utf8]{inputenc} 
\usepackage[T1]{fontenc}    
\usepackage{hyperref}       
\usepackage{url}            
\usepackage{booktabs}       
\usepackage{amsfonts}       
\usepackage{nicefrac}       
\usepackage{microtype}      
\usepackage{xcolor}         
\usepackage{graphicx}
\usepackage{multirow}
\usepackage{caption}

\usepackage{amsmath}

\title{FourierMamba: Fourier Learning Integration with State Space Models for Image Deraining}

\makeatletter
\def\thanks#1{\protected@xdef\@thanks{\@thanks
        \protect\footnotetext{#1}}}
\makeatother

\author{%
    Dong Li$^{\dagger}$, Yidi Liu$^{\dagger}$\thanks{{$^\dagger$} Co-first authors contributed equally, {$^\ast$} corresponding author.}, Xueyang Fu$^\ast$, Senyan Xu, Zheng-Jun Zha\\
    University of Science and Technology of China\\
}

\begin{document}

\maketitle

\begin{abstract}
Image deraining aims to remove rain streaks from rainy images and restore clear backgrounds. 
Currently, some research that employs the Fourier transform has proved to be effective for image deraining, due to it acting as an effective frequency prior for capturing rain streaks. 
However, despite there exists dependency of low frequency and high frequency in images, these Fourier-based methods rarely exploit the correlation of different frequencies for conjuncting their learning procedures, limiting the full utilization of frequency information for image deraining.  
Alternatively, the recently emerged Mamba technique depicts its effectiveness and efficiency for modeling correlation in various domains (e.g., spatial, temporal), and we argue that introducing Mamba into its unexplored Fourier spaces to correlate different frequencies would help improve image deraining.
This motivates us to propose a new framework termed FourierMamba, which performs image deraining with Mamba in the Fourier space.
Owning to the unique arrangement of frequency orders in Fourier space, the core of FourierMamba lies in the scanning encoding of different frequencies, where the low-high frequency order formats exhibit differently in the spatial dimension  (unarranged in axis) and channel dimension (arranged in axis).
Therefore, we design FourierMamba that correlates Fourier space information in the spatial and channel dimensions with distinct designs.
Specifically, in the spatial dimension Fourier space, we introduce the zigzag coding to scan the frequencies to rearrange the orders from low to high frequencies,  thereby orderly correlating the connections between frequencies; in the channel dimension  Fourier space with arranged orders of frequencies in axis, we can directly use Mamba to perform frequency correlation and improve the channel information representation.
Extensive experiments reveal that our method outperforms state-of-the-art methods both qualitatively and quantitatively.

\end{abstract}

\section{Introduction}
\label{Introduction}

Images taken in rainy conditions exhibit significant degradation in detail and contrast due to rain in the air, leading to unpleasant visual results and the loss of frequency information. This issue can severely impact the performance of outdoor computer vision systems, such as autonomous driving and video surveillance~\cite{wang2022online}. To mitigate the effects of rain, many image deraining methods~\cite{fu2011single,xiao2022image} have emerged in recent years, aiming to remove rain streaks and restore clear backgrounds in images.

\begin{figure}[t]
    \centering
    \includegraphics[width=\linewidth]{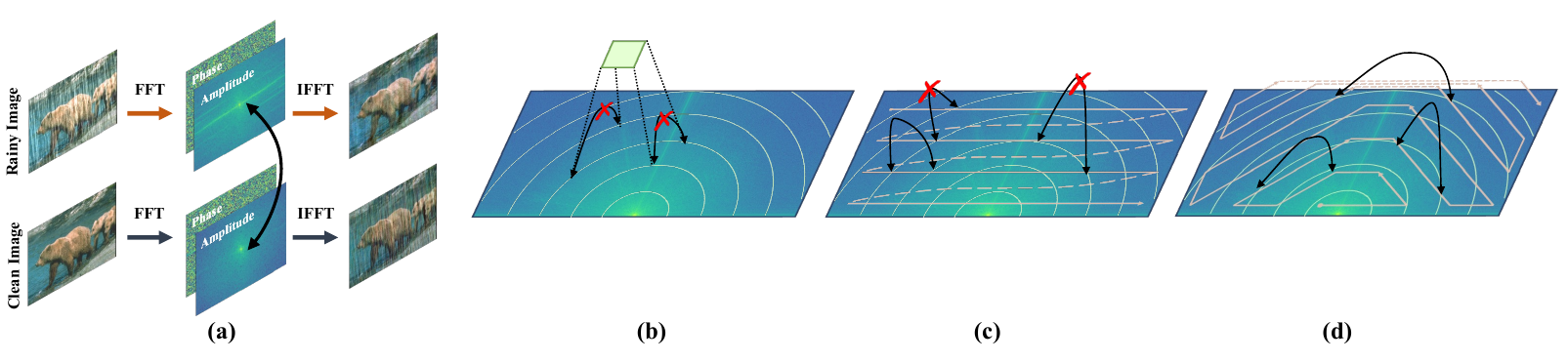}
    \caption{Observation and comparison of different frequency modeling methods. (a) Observation of the amplitude spectrum exchange. The degradation is mainly in amplitude components, so the Fourier transform helps to disentangle the image content and rain. (b) The commonly used $1\times1$ convolution cannot model the relationship between different frequencies. (c) Previous scanning in Fourier space will fail to establish the ordered dependence between frequencies. (d) Our proposed method achieves ordered frequency dependence from low to high (or vice versa), thus fully utilizing frequency information.}
    \vspace{-1.5em}
    \label{fig:intro}
\end{figure}

The advent of deep learning has spurred this field forward, with several learning-based deraining methods achieving remarkable success~\cite{fu2017removing,yang2017deep,zhang2018density}. Among them, some studies utilize the Fourier transform for deraining in the frequency domain~\cite{zhou2023fourmer,guo2022exploring}, proving effective. The key insights inspiring the use of the Fourier transform for image deraining are twofold: 1) The Fourier transform can separate image degradation and content components to some extent, serving as a prior for image deraining, as shown in Figure~\ref{fig:intro}; 2) The Fourier domain possesses global properties, where each pixel in Fourier space is involved with all spatial pixels. Thus, it makes sense to explore the task of rain removal using the Fourier transform. However, despite the existence of low frequency and high frequency dependencies in images, previous Fourier-based methods rarely utilize the correlation of different frequencies to combine their learning process. As shown in Figure~\ref{fig:intro}, the commonly used $1\times1$ convolutions cannot correlate different frequencies, limiting the full utilization of frequency information in the image. Therefore, we seek to exploit the beneficial properties of the Fourier transform while exploring correlating different frequencies.

Recently, an improved structured state-space sequence model (S4) with a selective scanning mechanism, Mamba, gives us hope. The selective methodology of Mamba can explicitly build correlation the correlation among image patches or pixels. Recent studies have witnessed the effectiveness and efficiency of Mamba in various domains such as spatial and temporal. Therefore, we believe that introducing Mamba into its unexplored Fourier space to correlate different frequencies will be advantageous for improving image deraining.

In this paper, we propose a novel framework named FourierMamba, which performs image deraining using mamba in the Fourier domain. Following the "spatial interaction + channel evolution" rule that has also been validated on Mamba~\cite{mambair,behrouz2024mambamixer}, we design the Mamba framework in the Fourier domain from both spatial and channel dimensions. Considering the unique arrangement of frequency orders in the Fourier domain, the core of FourierMamba lies in the scanning encoding of different frequencies, where the low-high frequency order formats unarranged in the spatial axis and arranged in the channel axis. Therefore, our proposed FourierMamba correlates Fourier space information in spatial and channel dimensions with distinct designs.

Specifically, \textbf{in the spatial dimension of the Fourier space}, low-high frequencies follow a concentric circular arrangement with lower frequencies near the center and higher frequencies around the periphery. If previous scanning method~\cite{vmamba} is used directly, the orderliness between frequencies will be destroyed, as shown in Figure~\ref{fig:intro}. We note that the zigzag coding in the JPEG compression field can place lower-frequency coefficients at the forefront of the array, while higher-frequency coefficients are positioned at the end. Hence, we introduce the zigzag coding to scan the frequency in the spatial dimension, rearranging the order from low to high frequency.  arrangement
Due to the symmetry of the frequency orders in the Fourier space, we do not directly employ the zigzag coding in its originally used space; instead, we implement it in a circling-like manner that matches the symmetric frequency orders in Fourier space.
In this way, this method orderly correlates the connections between frequencies, as shown in Figure~\ref{fig:intro}. 
\textbf{In the channel dimension of the Fourier space}, the frequency order is arranged along the axis, following the order of low in the middle to high on both sides. Therefore, we can directly use Mamba for frequency correlation, thus improving channel information representation and enhancing global properties on the channels.

In summary, our contributions are as follows: (1) We propose a novel framework FourierMamba that combines Fourier priors and State Space Model for correlating different frequencies in the Fourier space to enhance image deraining. (2) To rearrange the order from low to high frequency in the spatial dimension Fourier space, we propose a scanning method based on zigzag coding to orderly correlate different frequencies. (3) Based on the channel-dimension Fourier transform, we utilize Mamba to scan on the channels and correlate different frequencies to improve channel information representation. Extensive experiments demonstrate that the proposed FourierMamba surpasses state-of-the-art methods both qualitatively and quantitatively.


\section{Related Works}
\label{Related work}

\textbf{Image deraining.} Traditional image deraining methods focus on separating rain components by utilizing meticulously designed priors, such as Gaussian Mixture Models~\cite{li2016rain}, Sparse Representation Learning~\cite{gu2017joint,fu2011single}, and Directional Gradient Priors~\cite{ran2020single}. Although these methods are insightful, they often struggle to cope with complex precipitation patterns and the diverse real-world scenarios. The advent of deep learning has heralded a new era for image deraining. \cite{fu2017removing} introduces pioneering deep residual networks for image deraining. The initiation of CNNs marked a significant advancement, facilitating more nuanced and adaptive processing of rain streaks across a vast array of images~\cite{yang2017deep, zhang2018density}. With the evolution of transformers, the development of architectures that incorporate attention mechanisms~\cite{valanarasu2022transweather,wang2022uformer,xiao2022image} has further refined the capacity to recognize and eliminate rain components, addressing previous shortcomings in model generalization and detail preservation. In this work, we propose a novel baseline with a block based on Fourier and Mamba to enhance deraining performance.

\textbf{Fourier transform.}
Recently, the Fourier Transform has demonstrated its effectiveness in global modeling~\cite{chi2019fast,chi2020fast}. This transformation converts signals into a domain characterized by global statistical properties, facilitating advancements across various fields~\cite{huang2022deep,lee2018single,li2023embedding,pratt2017fcnn,xu2021fourier,yang2020fda}. Due to its efficacy in global modeling, the Fourier Transform has been introduced into low-level vision tasks~\cite{fuoli2021fourier,mao2023intriguing}. As an early attempt, \cite{fuoli2021fourier} proposes a Fourier Transform-based loss to optimize global high-frequency information for efficient image super-resolution. DeepRFT~\cite{mao2023intriguing} is proposed for image deblurring, employing a global receptive field to capture both low and high-frequency characteristics of various blurs, a concept similarly applied in image inpainting~\cite{suvorov2022resolution}. FECNet~\cite{huang2022deep} demonstrates that the amplitude of Fourier features decouples global luminance components, thereby proving effective for image enhancement. \cite{yu2022frequency} observes a similar phenomenon in image dehazing, where the amplitude reflects global haze-related information. In contrast, we introduce a progressive scanning strategy in the Fourier domain, enhancing the global modeling capability while addressing the directional sensitivity issues of visual Mamba.

\textbf{State Space Models.}
State Space Models (SSMs) have received a lot of attention recently due to their global modeling capabilities as well as linear complexity, with \cite{s4} initially introducing the base design of SSM models, and \cite{GSS} further enhancing their performance through gating units.More recently, the performance of Mamba \cite{mamba}, proposed based on selective scan mechanism and efficient hardware design, has seen significant enhancement. It stands as an efficient alternative to Transformers, finding applications in various domains including image classification \cite{vim}\cite{vmamba}, object detection\cite{mamba_in_mamba}, and remote sensing\cite{rsmamba}.In the field of image restoration, \cite{mambair} \cite{vmambair} initially introduced a general restoration framework based on the Mamba module but did not fully exploit the frequency domain information of images. \cite{freqmamba} introduced a wavelet transform branch, yet the scanning in the wavelet domain fails to fully extract global frequency domain information. This paper proposes a novel Mamba restoration network based on Fourier transform, aiming to comprehensively exploit the frequency domain information of images.

\begin{figure}[t]
    \centering
    \includegraphics[width=\linewidth]{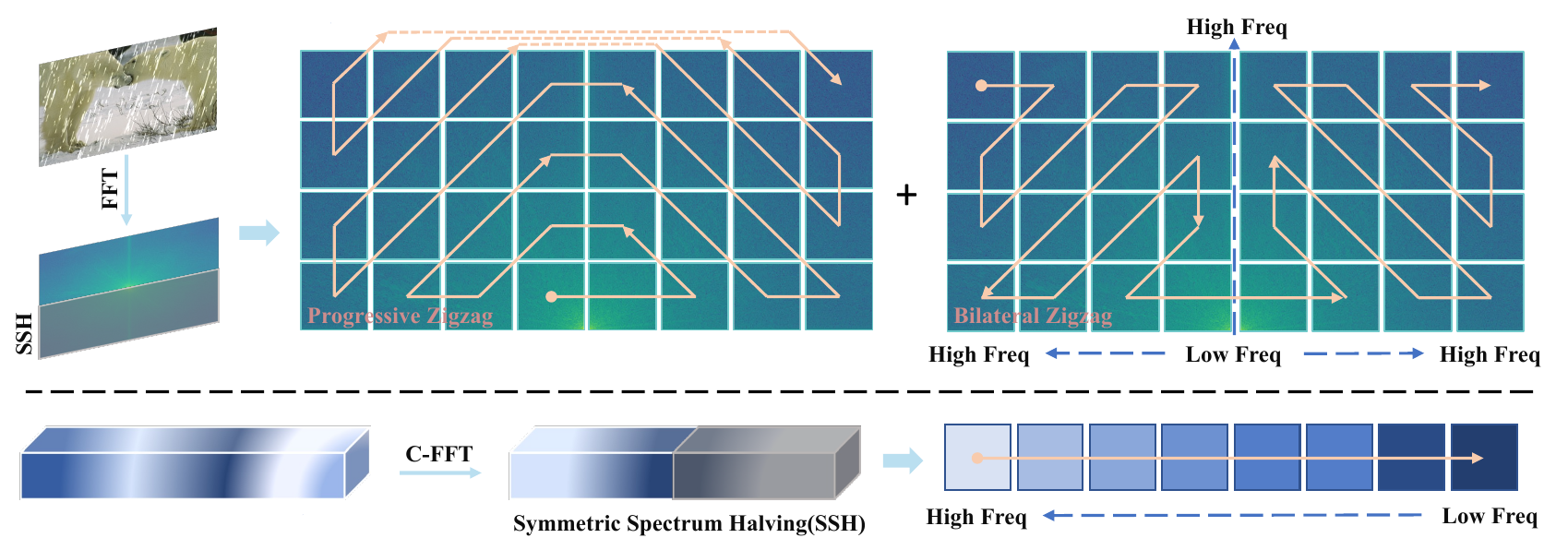}
    \caption{Our proposed Fourier space scanning method in the spatial dimension (top) and channel dimension (bottom). For simplicity, only one direction is shown for each scanning method, and in fact each method also performs a scan opposite to that shown.}
    \label{fig:scan}
\end{figure}

\subsection{Preliminary}
\textbf{Fourier transform.} Fourier transform is a widely used technique for analyzing the frequency content of an image. For images with multiple color channels, the Fourier transform is applied to each channel separately. Given an image $X\in \mathbb{R}^{H\times W\times C}$, the Fourier transform $\mathcal{F}$ converts it to Fourier space as the complex component F(x), which is expressed as:
\begin{equation}
\mathcal{F} \left ( x \right ) \left ( u,v \right ) = \frac{1}{\sqrt{HW} }\sum_{h=0}^{H-1}\sum_{w=0}^{W-1}x\left ( h,w \right )e^{- j2\pi\left (\frac{h}{H}u+  \frac{w}{W} v \right )} ,
  \label{eq:fft}
\end{equation}
where $u$ and $v$ indicate the coordinates of the Fourier space. $\mathcal{F}^{-1}\left ( x \right ) $ defines the inverse Fourier transform accordingly. Both the Fourier transform and its inverse procedure can be efficiently implemented using FFT/IFFT algorithms~\cite{frigo1998fftw}. The amplitude component $\mathcal{A}\left (x\right)\left ( u,v\right)$ and phase component $\mathcal{P}\left (x\right)\left ( u,v\right)$ are expressed as:
\begin{equation}
    \begin{split}           
  &\mathcal{A}\left (x\right)\left ( u,v\right)=\sqrt{R^{2}\left (x\right)\left (u,v\right)+I^{2}\left (x\right)\left (u,v\right)}, \\ 
  &\mathcal{P}\left (x\right)\left ( u,v\right)=\arctan \left [ \frac{I\left (x\right)\left (u,v\right)}{R\left (x\right)\left (u,v\right)} \right ],
    \end{split} 
  \label{eq:A-P}
\end{equation}
where $R(x)\left (u,v\right)$ and $I(x)\left (u,v\right)$ represent the real and imaginary parts respectively. The Fourier transform and its inverse procedure are applied independently to each channel of the feature maps.

\textbf{Channel-dimension Fourier transform.} 
We introduce the channel-dimension Fourier transform by individually applying Fourier transform along the channel dimension for each spatial position. For each position ($h\in \mathbb{R}^{H-1}$, $ w\in \mathbb{R}^{W-1}$) within $X\in \mathbb{R}^{H\times W\times C}$, denoted as $x(h,w,0: C-1)$ and abbreviated as $y(0: C-1)$, Fourier transform $\mathcal{F} \left ( \cdot \right )$ converts it to Fourier space as the complex component $\mathcal{F} \left ( y \right )$, which is expressed as:
\begin{equation}
\mathcal{F}(y(0: C-1))(z) = \frac{1}{C} \sum_{c=0}^{C-1}y(c)e^{- j2\pi\frac{c}{C}z } ,
  \label{eq:ch-fft}
\end{equation}.

Similarly, the amplitude component $\mathcal{A}(y(0: C-1))(z)$ and phase component $\mathcal{P}(y(0: C-1))(z)$ of $\mathcal{F}(y(0: C-1))(z)$ are expressed as:
\begin{equation}
\begin{split}
    &\mathcal{A}(y(0: C-1))(z) = \sqrt{R^{2}\left (y(0: C-1)\right)\left (z\right)+I^{2}\left (y(0: C-1)\right)\left (z\right)}, 
    \\ 
    &\mathcal{P}(y(0: C-1))(z) = \arctan \left [ \frac{I\left (y(0: C-1)\right)\left (z\right)}{R\left (y(0: C-1)\right)\left (z\right)} \right ].
\end{split}
\label{eq:CH-FFT_A-P}
\end{equation}
These operations can also be applied for the global vector derived by the pooling operation. In this way, $\mathcal{A}(z)$ and $\mathcal{P}(z)$ signify the magnitude and directional changes in the magnitude of various channel frequencies, respectively. Both of these metrics encapsulate global statistics related to channel information.

\textbf{State Space Models.} 
State Space Models (SSMs) serve as the cornerstone for transforming one-dimensional inputs into outputs through latent states, utilizing a framework of linear ordinary differential equations. Mathematically, SSMs can be formulated as follows, representing linear ordinary differential equations (ODEs):
\begin{equation}
    \begin{aligned} 
    h^{\prime}(t) & =\boldsymbol{A} h(t)+\boldsymbol{B} x(t), \\
    y(t) & =\boldsymbol{C} h(t)+\boldsymbol{D} x(t),
    \end{aligned}
\end{equation}
where, $h(t) \in \mathbb{R}^{N}$ denotes the hidden state vector, where N represents the size of the state. The parameters$\boldsymbol{A} \in \mathbb{R}^{N \times N}$, $\boldsymbol{B} \in \mathbb{R}^{N}$, and $\boldsymbol{C} \in \mathbb{R}^{N}$ are associated with the state size N, while 
 $\boldsymbol{D} \in \mathbb{R}^{1}$ represents the skip connection.
 
 Discrete versions of these models, such as Mamba\cite{mamba}, include a discretization step via the zero-order hold (ZOH) method. This enables the models to adaptively scan and adjust to the input data using a selective scanning mechanism. This mechanism provides a global receptive field with linear complexity, which is advantageous for image restoration tasks.

\begin{figure*}[t]
    \centering
    \includegraphics[width=0.8\linewidth]{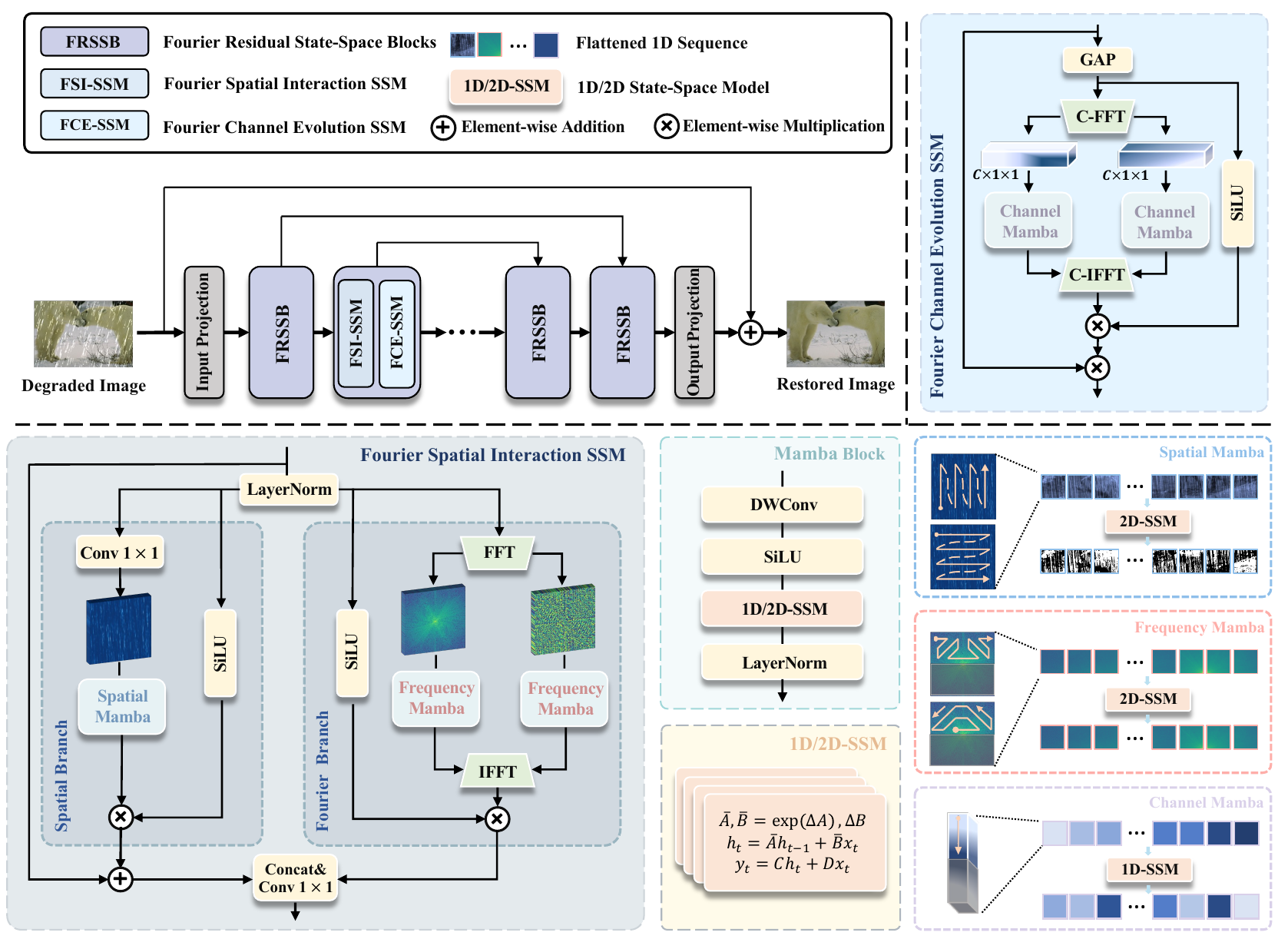}
    \caption{The overall architecture of the FourierMamba. Our FourierMamba consists of multiscale hierarchical design Fourier Residual State-Space Blocks(FRSSB). The core modules of FRSSB are Fourier Spatial Interaction SSM(FSI-SSM) and Fourier Channel Evolution SSM(FCS-SSM). }
    \vspace{-1.5em}
    \label{fig:architecture}
\end{figure*}

\subsection{Scanning in Fourier Space}
\label{sub:Scan}
Despite the unique characteristics of the selective scan mechanism (S6), it processes input data causally. Given the non-causal nature of visual data, directly applying this strategy to patches and flat images fails to estimate relations with unscanned patches, leading to a "directional sensitivity" issue constrained by the acceptance domain. Numerous methods have attempted to tackle this problem in the spatial domain~\cite{vmamba,mambair}. However, for image restoration, the Fourier space and its associated priors are crucial. Hence, we explore addressing the "directional sensitivity" issue within this domain. Specifically, we customize Fourier scanning strategies from both spatial and channel dimensions.

For the \textbf{spatial dimension}, each pixel point in the Fourier space contains global information, with its frequencies distributed in concentric circles. Scanning methods based on spatial arrangements~\cite{vmamba} disrupt the high-low frequency relationships in the frequency domain, thus hindering the modeling of image degradation information.

Therefore, we aim to devise a scanning method in the Fourier space to progressively model the frequency characteristics of images. An intuitive approach is to calculate the Euclidean distance from each point in the spectrum to the center point. On the shifted Fourier spectrum, the smaller the distance to the center point, the lower the frequency. The flaw of this intuitive approach is that for images of different sizes, it requires recalculating the Euclidean distance from each point to the center point. The additional computational overhead introduced by this flaw makes this approach impractical in the field of image restoration.

In JPEG compression, zigzag coding is commonly used among the Discrete Cosine Transform (DCT) coefficients of JPEG, where it prioritizes the energy-concentrated low-frequency coefficients at the beginning of the array, and places the less significant high-frequency coefficients towards the end, thereby facilitating more effective compression. Inspired by compression algorithms, we introduce a method that adopts the zigzag coding approach to scan the magnitude and phase spectra.

Additionally, due to the symmetry of the two-dimensional Fourier transform, scanning the entire spectrum would disrupt the symmetry in the Fourier space, potentially leading to the collapse of network optimization. Therefore, we scan half of the spectrum and then deduce the other half based on the central symmetry of the amplitude and the anti-central symmetry of the phase.

Specifically, we design two scanning strategies, as illustrated in
the Figure~\ref{fig:scan}. The first scanning method employs a dual zigzag pattern named \textbf{bilateral zigzag}, starting from the vertex of the highest frequency on one side of the spectrum, progressing in a zigzag pattern toward the center's low frequencies; similarly, it then zigzags to the opposite side's highest frequency. This scanning approach not only models the association between high and low frequencies but also takes into account the periodicity of the Fourier spectrum. Due to the periodic nature of the Fourier transform, the high-frequency ends on either side should, in fact, be contiguous. The second method builds upon the low-to-high frequency sequence established by zigzag scanning and conducts a scan from low to high frequencies, which is named \textbf{progressive zigzag}. This method is motivated by the tendency of neural networks to initially learn low-frequency information when extracting image characteristics. Following the previous method \cite{vmamba,mambair}, we reverse the above two scanning methods as additional scanning directions.

For the \textbf{channel dimension} Fourier space, since it is a one-dimensional sequence arranged in order of low to high frequencies, we directly scan it one-dimensionally. Similarly, due to  the symmetry of the Fourier transform, we scan only half and derive the other half. Through Fourier space scanning in both spatial and channel dimensions, we can correlate the connections between frequencies in an orderly manner, thereby making full use of frequency information to improve rain removal.

\begin{figure}
    \centering
    \includegraphics[width=1\linewidth]{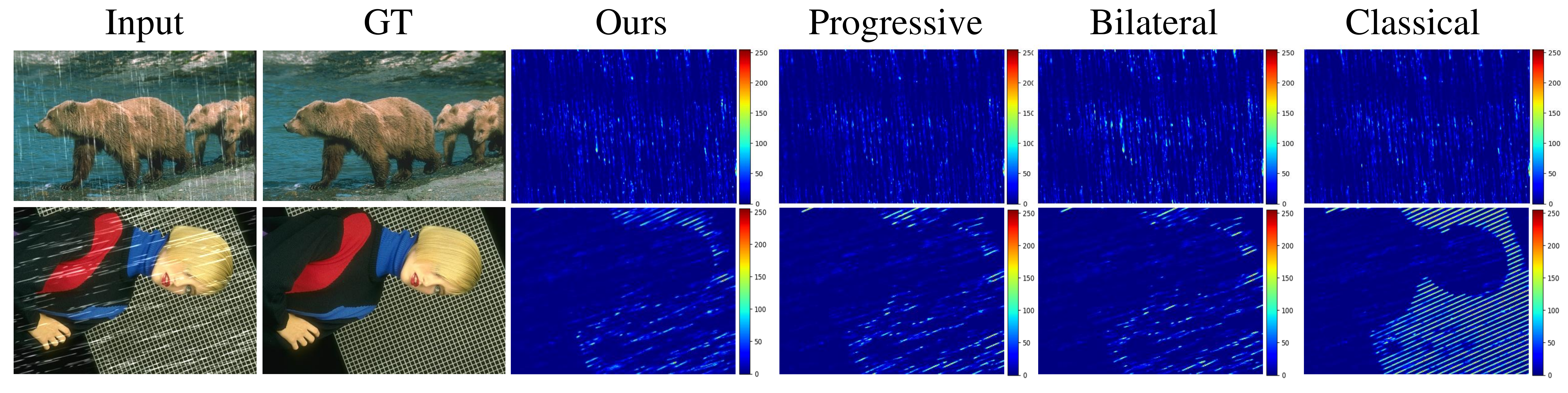}
    \caption{The error map between the GT and the restored images using various scanning methods in Fourier space. The two scanning methods we propose can achieve smaller errors than using classical scanning method~\cite{vmamba}. And the combination of the two scanning methods is better than either one.}
    \vspace{-1.8em}
    \label{fig:error map scanning}
\end{figure}

\subsection{FourierMamba}

\subsubsection{Overall Framework}

In Figure~\ref{fig:architecture}, we illustrate our proposed FourierMamba. Given a rainy image $I \in \mathbb{R}^{H\times W\times 3}$, FourierMamba first uses $3\times3$ convolution layers to generate shallow features with dimensions of $H\times W\times C$, where $H$ and $W$ represent height and width, and $C$ denotes the number of channels. Subsequently, we employ a multi-scale U-Net architecture to obtain deep features. This stage consists of a stack of Fourier Residual State-Space Groups, each containing several Fourier Residual State-Space Blocks (FRSSB). The FRSSB incorporates our two core designs: the Fourier Spatial Interaction SSM block and the Fourier Channel Evolution SSM block. They correlate Fourier domain information from spatial and channel dimensions, respectively, to fully leverage frequency information.

\subsubsection{Fourier Spatial Interaction SSM}

The structure of the Fourier Spatial Interaction State Space Model (FSI-SSM) is shown in Figure~\ref{fig:architecture}. We first apply LayerNorm to transform the input features $F_{in}$ into $F_l$. To facilitate the interaction between spatial and frequency information, FSI-SSM employs both a Fourier branch and a spatial branch to collaboratively process $F_{in}$.

\textbf{Fourier Branch:} $F_l$ is transformed into the Fourier spectrum through the Fast Fourier Transform, subsequently decomposed into the amplitude spectrum $\mathcal{A}(F_l)$ and phase spectrum $\mathcal{P}(F_l)$.
The amplitude spectrum and phase spectrum are then processed separately using the progressive frequency scanning method illustrated in Figure~\ref{fig:scan} to obtain $\mathcal{A}'(F_l)$ and $\mathcal{P}'(F_l)$.
\begin{equation}
\begin{split}  
&\mathcal{A}'(F_l)={\rm FourScan}(\mathcal{A}(F_l)),\\
&\mathcal{P}'(F_l)={\rm FourScan}(\mathcal{P}(F_l)),
\end{split}
\end{equation}
where ${\rm FourScan}$ is the sequence transformation using the Fourier space scan described in Sec.~\ref{sub:Scan}. Following a series of works~\cite{vmamba,mambair,freqmamba}, the sequence transformation employs the following operation sequence: $DWConv\to SiLU\to SSM \to LayerNorm$.
We then perform an inverse Fourier transform on the processed spectrum and multiply it with the output of $\rm SiLU$.
\begin{equation}
F_f = (\mathcal{F}^{-1}(\mathcal{A}'(F_l),\mathcal{P}'(F_l)))\odot {\rm SiLU}(F_l),
\end{equation}
where $F_f$ is the output of the fourier branch, and $\odot$ is the Hadamard product.

\textbf{Spatial Branch}
In the spatial domain,  we feed the input features $F_l$ into two parallel sub-branches. One sub-branch activates the features using the $\rm SiLU$ function. The other sub-branch performs spatial Mamba on features after $1\times1$ convolution. Specifically, spatial Mamba adopts the same operation sequence as the above frequency branch but the scanning in SSM uses the two-dimensional selective scanning module shown in Figure~\ref{fig:architecture}, which follows previous work~\cite{vmamba,mambair}. Finally, the outputs of the two sub-branches are multiplied element-wise to obtain the output $F_s$.
\begin{equation}
F_s = {\rm SpaScan}({\rm Conv}(F_l))\odot {\rm SiLU}(F_l),
\end{equation}
where ${\rm Conv}$ is $1\times1$ convolution and {\rm SpaScan} is the spatial Mamba mentioned above.
Subsequently, we employ a residual connection to add the spatial output to $F_{in}$. The spatial branch captures global features in the spatial domain which complement the frequency correlations captured by the Fourier branch in the frequency domain, thereby benefiting the performance of image deraining. Hence, we concatenate the outputs of the spatial and frequency branches and use a $1\times1$ convolution for the fusion of spatial and frequency information.

\subsubsection{Fourier Channel Evolution SSM}

Previous work~\cite{mambair} claims that selecting key channels can avoid channel redundancy in SSM. Since each channel contains the information of all channels after the channel-dimension Fourier transform, we perform channel interaction in the Fourier domain to efficiently correlate different frequencies of channels. As depicted in Figure~\ref{fig:architecture}, our proposed Fourier Channel Evolution SSM (FCS-SSM) consists of three sequential parts: applying the Fourier transform along the channel dimension to obtain channel-wise Fourier domain features, scanning its amplitude and phase, then restoring to the spatial domain. Specifically, assuming the input features are denoted as $F_r \in \mathbb{R}^{H_r\times W_r\times C_r}$, we first perform global average pooling on it.
\begin{equation}
F_g = \frac{1}{H_rW_r}\sum_{h=0}^{H_r-1} \sum_{w=0}^{W_r-1}F_g(h,w) ,
  \label{GAP}
\end{equation}
where $F_g \in \mathbb{R}^{1\times 1\times C_r}$ corresponds to the center point of the amplitude spectrum of $F_r$ (see supplementary material), which effectively encapsulates the global information of the feature. Then, we use the channel-dimensional Fourier transform shown in Equ.~\ref{eq:ch-fft} on $F_g$ to obtain $\mathcal{F}(F_g)(z)$. Based on this, we use Equ.~\ref{eq:CH-FFT_A-P} for $\mathcal{F}(F_g)(z)$ to obtain its amplitude component $\mathcal{A}(F_g)(z)$ and phase component $\mathcal{P}(F_g)(z)$. Since the amplitude spectrum and phase spectrum have obvious information meaning, we choose to perform Mamba scanning on these two components.
\begin{equation}
\begin{split}  
&\mathcal{A}(F_g)(z)'={\rm ChaScan}(\mathcal{A}(F_g)(z)),\\
&\mathcal{P}(F_g)(z)'={\rm ChaScan}(\mathcal{P}(F_g)(z)),
\end{split}
\end{equation}
where ${\rm ChaScan}$ is a one-dimensional sequence transformation that uses the following sequence of operations: $DWConv\to SiLU\to SSM\to LayerNorm$. Its scanning method is shown in Figure~\ref{fig:scan}. After the Mamba correlates different frequencies in the channel dimension, we perform an inverse Fourier transform on it and multiply the result with the channel features after $\rm SiLU$ activation.
\begin{equation}
F_a = (\mathcal{F}^{-1}(\mathcal{A}(F_g)(z)',\mathcal{P}(F_g)(z)'))\odot {\rm SiLU}(F_g),
\end{equation}
where $F_a \in \mathbb{R}^{1\times 1\times C_r}$ is the channel feature after correlating different frequencies. Finally, we multiply it with the spatial feature in a form of attention to get the output $F_c\in \mathbb{R}^{H_r\times W_r\times C_r}$.
\begin{equation}
F_c = F_a\odot F_r.
\end{equation}

\subsubsection{Optimization}

We impose constraints in both the spatial and frequency domains. In the spatial domain, we utilize the L1 loss between the final output $ Y_{out}$ and the ground truth $Y_{gt}$. In the frequency domain, we apply the L1 loss based on the Fourier transform. The overall loss function is formulated as follows:
\begin{equation}
{\mathcal{ L}_{total}} = \left \| Y_{out}-Y_{gt} \right \|_1 +\lambda\left \|\mathcal{F}(Y_{out}) - \mathcal{F}(Y_{gt})\right \|_1,
  \label{loss}
\end{equation}
where $\lambda$ is the balancing weight. In particular,$\lambda$ is set to 0.02 empirically.

\section{Experiment}
\label{Experiment}

\subsection{Experimental Settings}
\label{Experimental Settings}

\textbf{Datasets.}
For training, we employ the widely used Rain13k dataset~\cite{chen2021hinet}.  It contains 13,712 image pairs in the training set, and we evaluate the results on Rain100H~\cite{yang2017deep}, Rain100L~\cite{yang2017deep}, Test2800~\cite{fu2017removing}, and Test1200~\cite{zhang2018density}.

\begin{figure}
    \centering
    \includegraphics[width=0.9\linewidth]{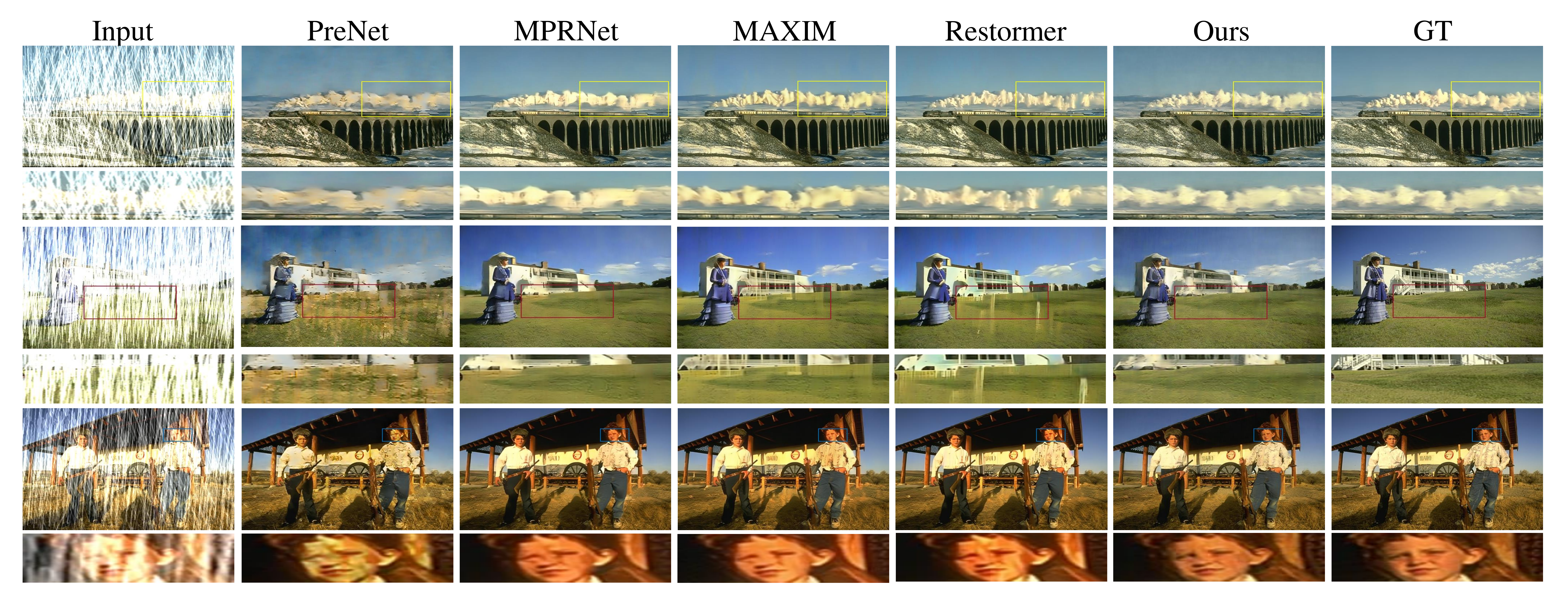}
    \caption{Visual quality comparison on Rain100H~\cite{yang2017deep}. Zoom in for better visualization.}
    \vspace{-1.8em}
    \label{fig:visual rain100H}    
\end{figure}

\textbf{Implementation details.}
Our model is implemented within the PyTorch framework and executed on an NVIDIA A100 GPU. The number of blocks per layer has an impact on both the model's parameter count and its deraining performance. After balancing the weights, we configure the blocks per layer as [2, 3, 3, 4, 3, 3, 2], which allows us to achieve commendable performance with a reasonable number of parameters. We adopt the progressive training strategy. Specifically, we set the total number of iterations to 80,000 and image sizes to [160, 256, 320, 384], with the corresponding batch sizes of [8, 4, 2, 1]. We utilize the Adam optimizer with default parameters. The initial learning rate is established at $3 \times e^{-4}$, followed by a gradual decay to $1 \times e^{-6}$ using a cosine annealing schedule.

\subsection{Comparison with State-of-the-art Methods}

We compare our method with state-of-the-art deraining methods: DerainNet \cite{Fu_2017}, UMRL \cite{Yasarla_2019_CVPR}, RESCAN \cite{li2018recurrent}, PreNet \cite{ren2019progressive}, MSPFN \cite{Kui_2020_CVPR} , SPAIR \cite{purohit2021spatially} , MPRNet \cite{Zamir2021MPRNet} , Restormer\cite{Zamir2021Restormer}, Fourmer \cite{zhou2023fourmer}, IR-SDE \cite{luo2023refusion}, MambaIR \cite{mambair} VMambaIR\cite{vmambair} and FreqMamba\cite{freqmamba}.

\begin{table*}
  \centering
  \caption{Quantitative comparison (PSNR/SSIM) for Image Deraining on five benchmark datasets. The highest and second-highest performances are marked in bold and underlined. '-' indicates the result is not available.}
  \vspace{-2mm}
   \resizebox{1\linewidth}{!}{
    \begin{tabular}{c|c|cccccccc|c c}
    \toprule[0.1em]
    \multirow{2}{*}{\textbf{Method}}&\multirow{2}{*}{\textbf{Venue}}& \multicolumn{2}{c}{\textbf{Rain100H}~\cite{yang2017deep}} & \multicolumn{2}{c}{\textbf{Rain100L}~\cite{yang2017deep}} & \multicolumn{2}{c}{\textbf{Test2800}~\cite{fu2017removing}} & \multicolumn{2}{c|}{\textbf{Test1200}~\cite{zhang2018density}} &\multirow{2}{*}{\textbf{Param(M)}}&\multirow{2}{*}{\textbf{GFlops}}\\
         &   & PNSR $\uparrow$& SSIM $\uparrow$& PNSR $\uparrow$& SSIM $\uparrow$& PNSR $\uparrow$& SSIM $\uparrow$& PNSR $\uparrow$& SSIM $\uparrow$ & &\\
    \midrule
    DerainNet~\cite{fu2017removing}  &TIP'17& 14.92 & 0.592 & 27.03 & 0.884 & 24.31 & 0.861 & 23.38 & 0.835 & 0.058 & 1.453\\
    UMRL~\cite{Yasarla_2019_CVPR} &CVPR'19 &  26.01 & 0.832 & 29.18 & 0.923 & 29.97 & 0.905 & 30.55 & 0.910  & 0.98 & -\\
    RESCAN~\cite{li2018recurrent}  &ECCV'18& 26.36 & 0.786 & 29.80 & 0.881 & 31.29 & 0.904 & 30.51 & 0.882&1.04 &20.361\\
    PreNet~\cite{ren2019progressive} &CVPR'19& 26.77 & 0.858 & 32.44 & 0.950 & 31.75 & 0.916 & 31.36 & 0.911&0.17&73.021\\
    MSPFN~\cite{Kui_2020_CVPR}  &CVPR'20 & 28.66 &  0.860 &  32.40 & 0.933 & 32.82 & 0.930 & 32.39 & 0.916  &13.22 &604.70\\
    SPAIR~\cite{purohit2021spatially}  &ICCV'21 & 30.95 & 0.892 & 36.93 & 0.969 & 33.34 & 0.936 & 33.04 & 0.922&-&-\\
    MPRNet~\cite{Zamir2021MPRNet} &CVPR'21& 30.41 & 0.890 & 36.40 & 0.965 & 33.64 & 0.938 & 32.91 & 0.916&3.64&141.28\\
    
    Restormer~\cite{Zamir2021Restormer}  &CVPR'22 & 31.46 &   0.904 & {38.99} & 0.978 & {34.18} & {0.944} & {33.19} & {0.926} &24.53&174.7\\
    Fourmer~\cite{zhou2023fourmer}  &ICML'23 & 30.76 &   0.896 & 37.47 & 0.970 & - & - & 33.05 & 0.921 &0.4&16.753\\
    IR-SDE~\cite{luo2023image}  &ICML'23 & 31.65 &   0.904 & 38.30 & {0.980} & 30.42 & 0.891 & - & - &135.3&119.1\\
    MambaIR~\cite{mambair}  & arxiv'24 & 30.62 &   0.893 & 38.78 & 0.977 & 33.58 & 0.927 & 32.56 & 0.923 &31.51&80.64\\
    VMambaIR~\cite{vmambair}  &arxiv'24 & {31.66} &   {0.909} & {39.09} & 0.979 & 34.01 & 0.944 & {33.33} & 0.926 &-
    &-\\
    FreqMamba~\cite{freqmamba} & arxiv'24 & \underline{31.74} & \underline{0.912} & \underline{39.18} & \underline{0.981} & \textbf{{34.25}} & \textbf{{0.951}}  & \underline{33.36} & \underline{0.931} &14.52&36.49\\
    \midrule[0.1em]
    FourierMamba(Ours) & - & \textbf{31.79} & \textbf{0.913} & \textbf{39.73} & \textbf{0.986} & \underline{34.23} & \underline{0.949}  & \textbf{34.76} & \textbf{0.938} &17.62&22.56\\
    \bottomrule[0.1em]
    \end{tabular}%
    }
  \label{tab:derain_qua}%
  \vspace{-2mm}
\end{table*}%

\textbf{Quantitative Comparison.} Following previous work, we compute the Peak Signal-to-Noise Ratio (PSNR) and Structural Similarity (SSIM) scores using the Y channel in the YCbCr color space. Table~\ref{tab:derain_qua} reports the performance evaluation on four datasets. It can be seen that our method achieves the highest PSNR and SSIM across all datasets, which underscores the efficacy of FourierMamba in enhancing deraining performance.

\textbf{Qualitative Comparison.}
To demonstrate the enhanced fidelity and detail levels exhibited by the images generated by our proposed FourierMamba, we compare the visual quality of challenging degraded images from the Rain100H dataset in Figure~\ref{fig:visual rain100H}. Our method achieves excellent results when faced with complex or extremely severe rain streaks. Compared to previous methods, our FourierMamba achieves impeccable performance in both global and local restoration. For instance, by zooming into the red boxed area in Figure~\ref{fig:visual rain100H}, our method removes more rain streak residues while better restoring texture details. We provide additional visual results in the Appendix.

\subsection{Ablation Studies}

We perform ablations on the key designs and scanning methods of the framework on the Rain100L.

\textbf{Fourier Spatial Interaction SSM (FSI-SSM) and Fourier Channel Evolution SSM (FCI-SSM).} We replace the mamba scan in FSI-SSM and FCE-SSM with $1\times1$ convolution, called w/o FSI-SSM and w/o FCE-SSM respectively. It can be seen from Table~\ref{tab:ablation_block} that since $1\times1$ convolution cannot model the dependence of different frequencies, its performance is worse than the mamba scan in the Fourier domain in both the spatial dimension and the channel dimension.

\textbf{Fourier prior.} We do not use Fourier transform in the spatial dimension and channel dimension respectively, but directly perform mamba scanning, which are called without spatial dimension Fourier (w/o SDF) and without channel dimension Fourier (w/o CDF) respectively. It can be seen from Table~\ref{tab:ablation_block} that after losing the Fourier prior in the spatial dimension and channel dimension, the performance drops significantly. This proves the effectiveness of Fourier prior for removing rain from images. 

\textbf{Scanning method in Fourier space.} We compare several scanning methods of the spatial dimension Fourier space, with the same amount of calculation. Table~\ref{tab:scan} illustrates that the performance of the two scanning methods we proposed is better than the classic two-dimensional scanning method~\cite{vmamba}. And thanks to complementarity, the combination of the two methods can also further improve performance. The visual comparison in Figure~\ref{fig:error map scanning} supports this.

\vspace{-6mm}
\begin{table}[]
    \centering
    \caption{Ablation studies of key designs in the proposed method.}
    \begin{tabular}{cccccc}
        \hline
          & w/o FSI-SSM &w/o FCE-SSM & w/o SDF   & w/o CDF &  Ours
        \\
         PSNR  &39.05 & 39.08  & 38.25  &  38.72 & \textbf{39.73}
        \\
        SSIM    & 0.9835  & 0.9836 &  0.9810  &  0.9827 & \textbf{0.9856}
        \\
        \hline
        
    \end{tabular}
    \vspace{-1em}
    \label{tab:ablation_block}
\end{table}

\begin{table}[ht]
    \centering
    \caption{Ablation study of different scanning methods in Fourier space.}
    \begin{tabular}{ccccc}
    \hline
            & Classic\cite{vmamba} & Bilateral & Progressive     & Ours\\
       PSNR  &38.82 &39.31  &39.28    & \textbf{39.73} \\
         SSIM  &0.9817 &0.9844  &0.9843  & \textbf{0.9856}
         \\
    \hline
    \end{tabular}
    \vspace{-2em}
    \label{tab:scan}
\end{table}

\section{Conclusion}
In this paper, we propose a novel image deraining framework, FourierMamba, which utilizes mamba to correlate frequencies in the Fourier space, thus fully exploiting frequency information. Specifically, we design the mamba framework by integrating the unique arrangement of frequency orderings within the Fourier domain across spatial and channel dimensions. In the spatial dimension, we devise two zigzag-based methods to scan frequencies, systematically correlating them. In the channel dimension, due to the ordered arrangement of frequencies along the axis, we directly apply mamba for frequency correlation. This work introduces a new research strategy to address the underutilization of frequency information in image deraining that affects performance. Extensive experimental results on multiple benchmarks validate the effectiveness of the proposed method.

\begin{figure*}[!t]
    \centering
    \begin{minipage}[b]{0.48\textwidth}
        \centering
        \includegraphics[width=\textwidth]{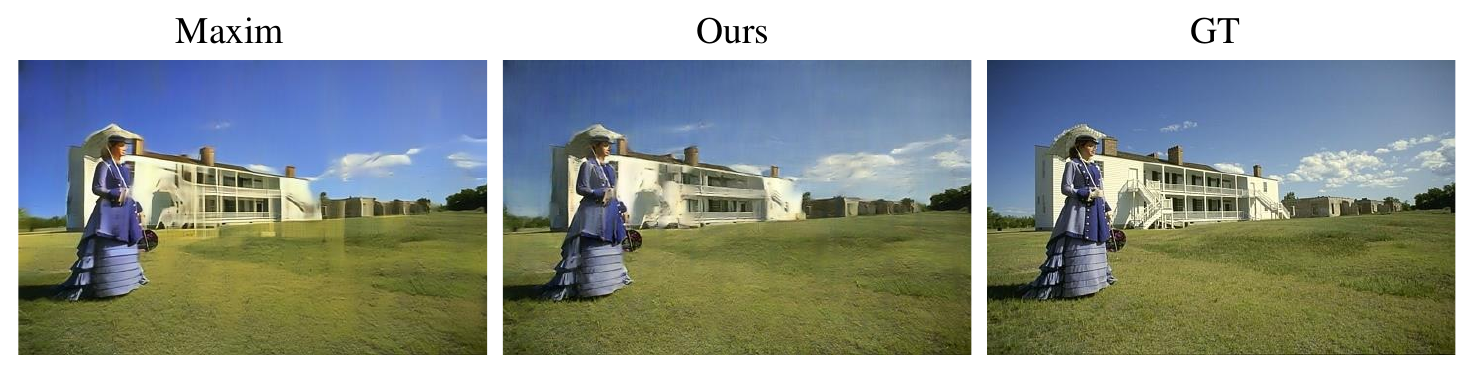}
        \caption{Correction of the second picture in Figure 5.}
    \end{minipage}
    \hfill
    \begin{minipage}[b]{0.48\textwidth}
        \centering
        \includegraphics[width=\textwidth]{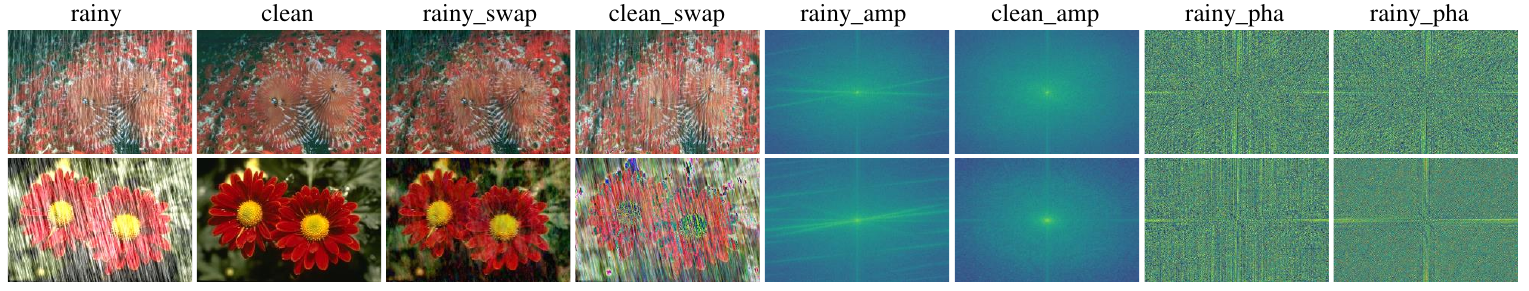}
        \caption{More cases of the observation. Rainy\_swap and clean\_swap represent the rainy and clean images after swapping their amplitude spectra, respectively.}
    \end{minipage}
    \vspace{0.5cm}
    \begin{minipage}[b]{0.48\textwidth}
        \centering
        \includegraphics[width=\textwidth]{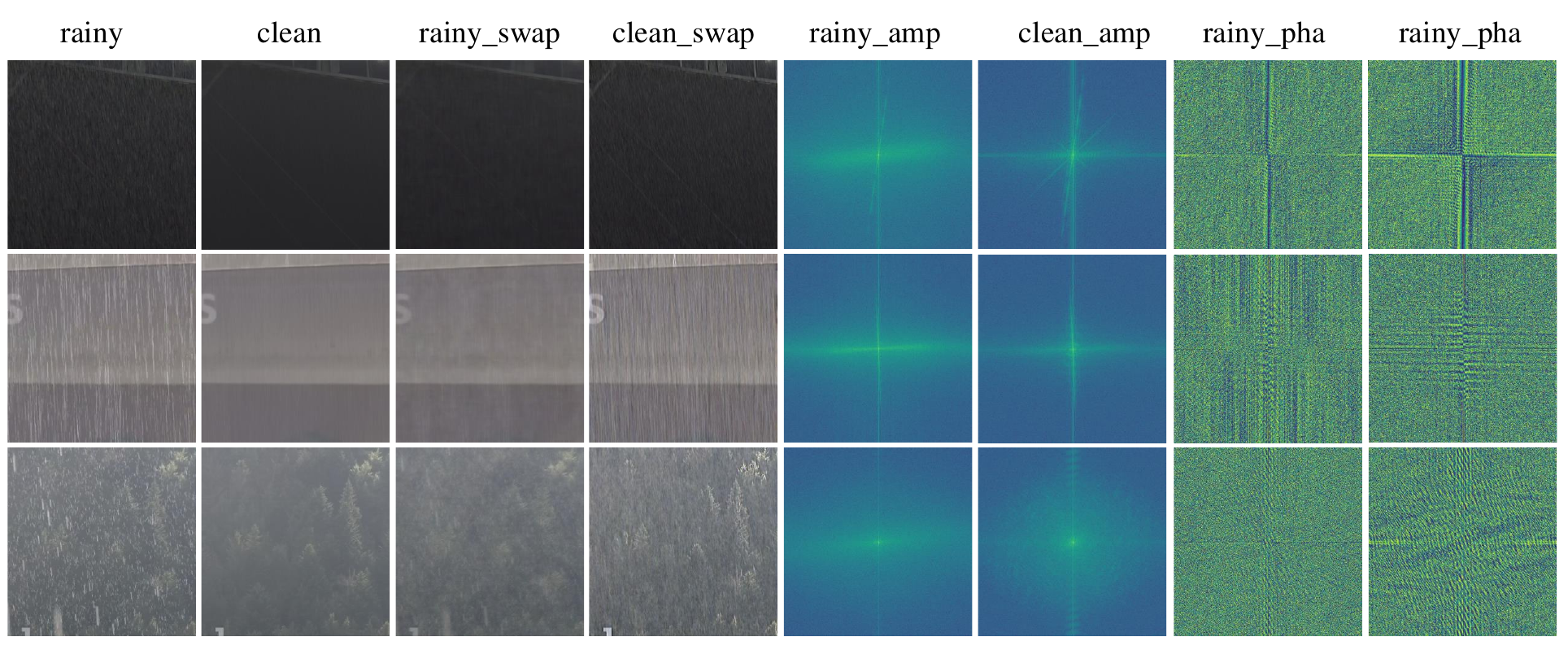}
        \caption{The observation on real-world images. They also align with our conclusion.}
    \end{minipage}
    \hfill
    \begin{minipage}[b]{0.48\textwidth}
        \centering
        \includegraphics[width=\textwidth]{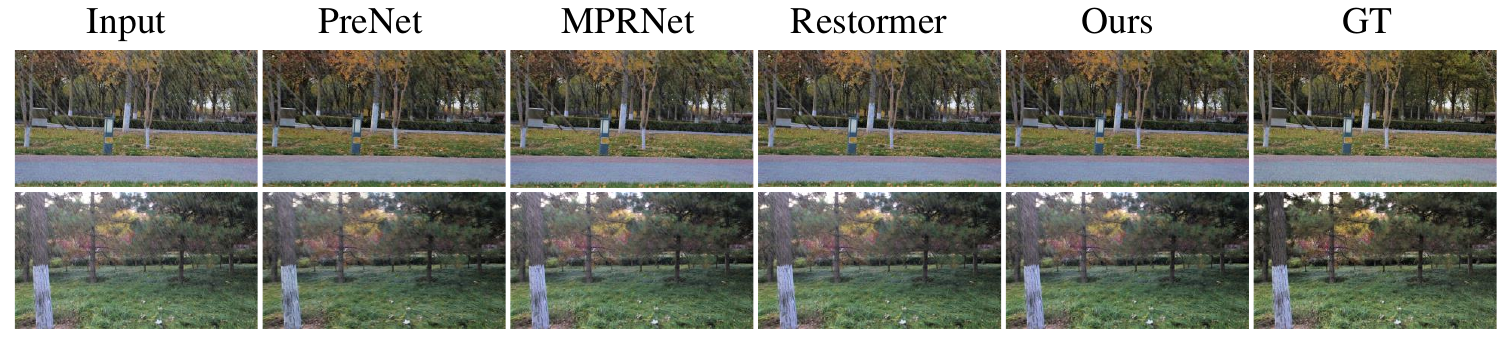}
        \caption{Qualitative comparison of real-world rainy images from RainDS-Real.}
    \end{minipage}
    \vspace{0.5cm}
    \begin{minipage}[b]{0.48\textwidth}
        \centering
        \includegraphics[width=\textwidth]{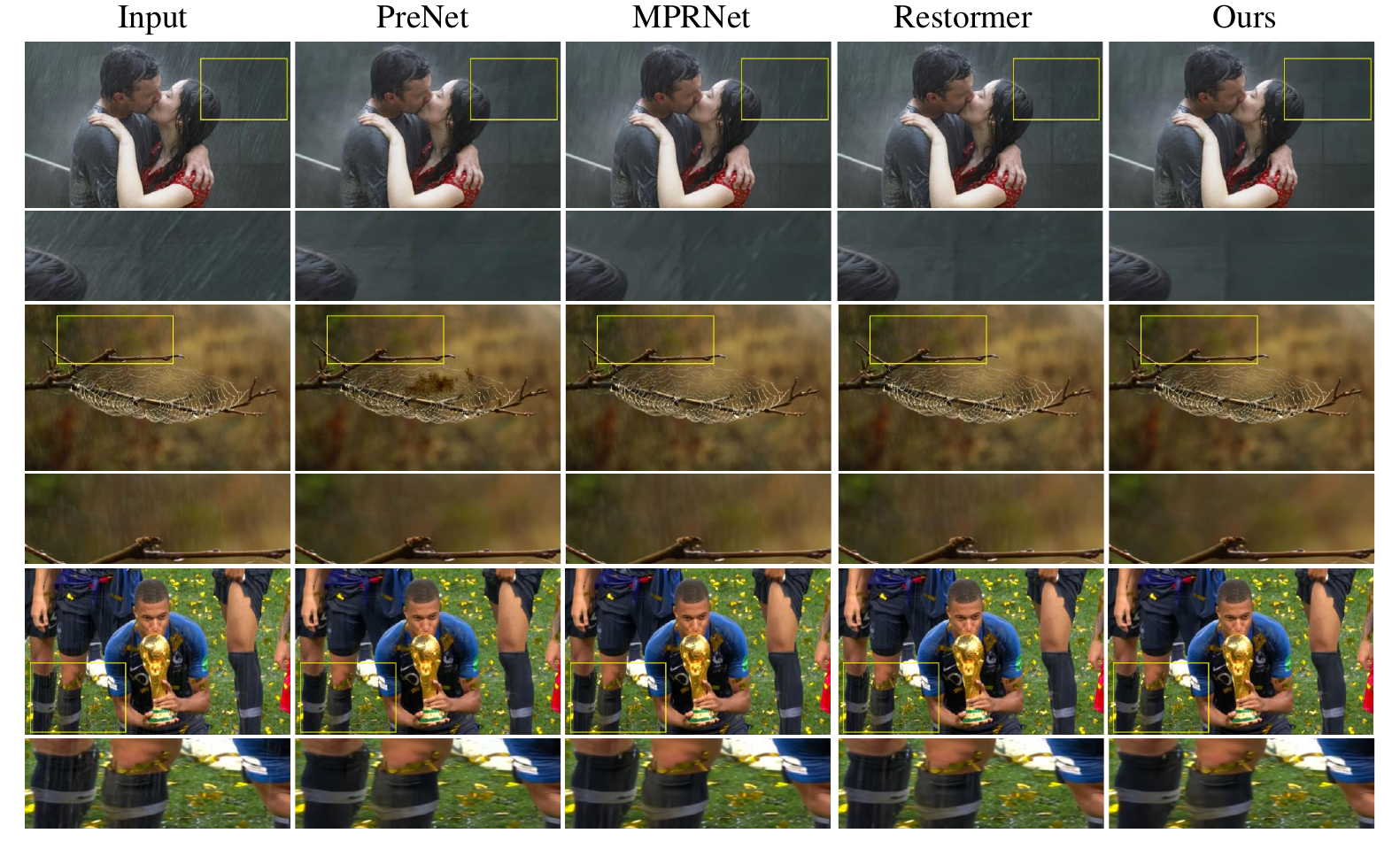}
        \caption{Qualitative comparison of real-world rainy images from Internet-Data.}
    \end{minipage}
    \hfill
    \begin{minipage}[b]{0.48\textwidth}
        \centering
        \includegraphics[width=\textwidth]{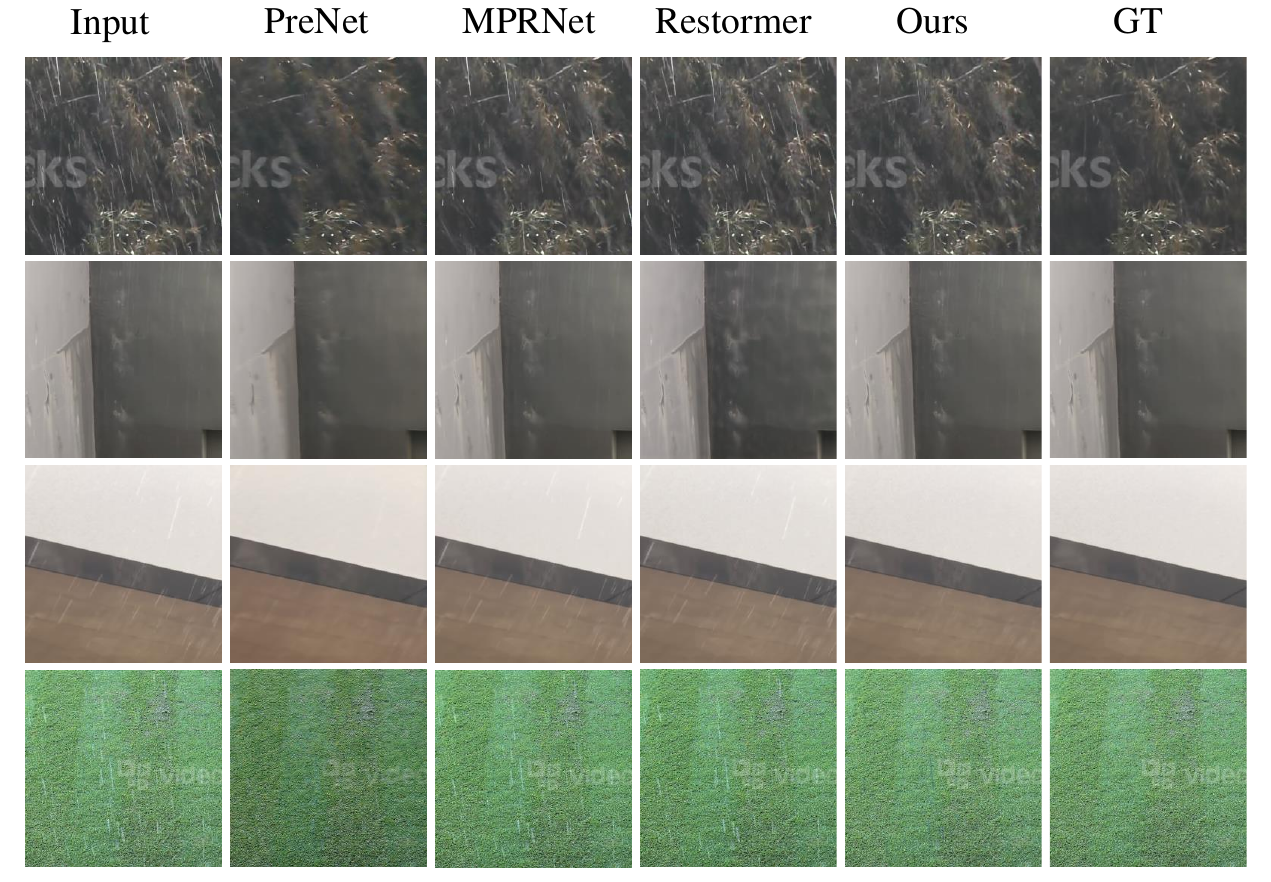}
        \caption{Qualitative comparison of real-world rainy images from SPA-Data.}
    \end{minipage}
    \vspace{0.5cm}
    \begin{minipage}[b]{0.48\textwidth}
        \centering
        \includegraphics[width=\textwidth]{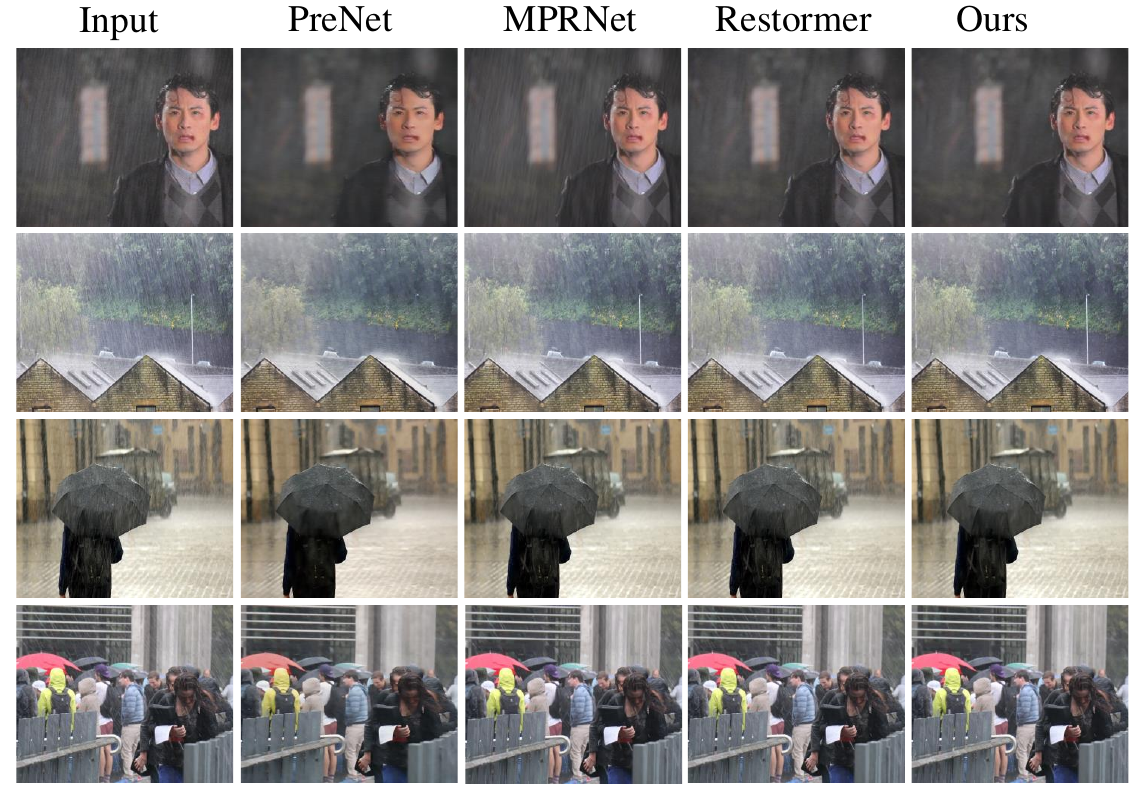}
        \caption{Qualitative comparison of real-world rainy images from RE-RAIN.}
    \end{minipage}
\end{figure*}


\clearpage



\medskip
{
\bibliography{neurips_2024}
}







\end{document}